\documentclass[11pt]{article}
\usepackage[a4paper]{geometry}
\usepackage[font=footnotesize,labelfont=bf]{caption}
\usepackage{fancyhdr}
\usepackage{subcaption}
\usepackage{placeins}
\usepackage{amsmath}
\usepackage{amsfonts}
\usepackage{graphicx,psfrag,epsf}
\usepackage{enumerate}
\usepackage[linesnumbered,ruled]{algorithm2e}
\usepackage{tabu}
\usepackage{setspace}
\usepackage{tikz}
\usepackage{float}
\usepackage{bm}
\usepackage{color}
\usepackage{listings}
\usepackage[square,sort,comma,numbers]{natbib}
\usepackage[hidelinks]{hyperref}
\usepackage[open]{bookmark}
\usepackage{booktabs}
\usepackage{rotating}
\usepackage{authblk}

\DeclareMathOperator*{\argmax}{arg\,max}
\SetArgSty{textnormal}
\SetKwInput{KwTrain}{Train}
\newtheorem{lemma}{Lemma}
\newtheorem{corollary}{Corollary}[lemma]

\fancypagestyle{firststyle}
{
	\fancyhf{}
	\fancyfoot[L]{\fontsize{9}{11}\selectfont Accepted for publication in \textit{Information Fusion} at https://doi.org/10.1016/j.inffus.2020.03.007. \newline \copyright \ 2020. This manuscript version is made available under the CC-BY-NC-ND 4.0 license http://creativecommons.org/licenses/by-nc-nd/4.0/}
}

\begin{document}

\title{Stacked Penalized Logistic Regression for Selecting Views in Multi-View Learning}
\date{February 6, 2020}
\author[1]{Wouter van Loon}
\author[1]{Marjolein Fokkema}
\author[2]{Botond Szabo}
\author[1]{Mark de Rooij}
\affil[1]{Department of Methodology and Statistics, Leiden University}
\affil[2]{Mathematical Institute, Leiden University}
\maketitle

\thispagestyle{firststyle}

\begin{abstract}
	In biomedical research, many different types of patient data can be collected, such as various types of omics data and medical imaging modalities. Applying multi-view learning to these different sources of information can increase the accuracy of medical classification models compared with single-view procedures. However, collecting biomedical data can be expensive and/or burdening for patients, so that it is important to reduce the amount of required data collection. It is therefore necessary to develop multi-view learning methods which can accurately identify those views that are most important for prediction. \par
	In recent years, several biomedical studies have used an approach known as multi-view stacking (MVS), where a model is trained on each view separately and the resulting predictions are combined through stacking. In these studies, MVS has been shown to increase classification accuracy. However, the MVS framework can also be used for selecting a subset of important views. \par
	To study the view selection potential of MVS, we develop a special case called stacked penalized logistic regression (StaPLR). Compared with existing view-selection methods, StaPLR can make use of faster optimization algorithms and is easily parallelized. We show that nonnegativity constraints on the parameters of the function which combines the views play an important role in preventing unimportant views from entering the model. We investigate the performance of StaPLR through simulations, and consider two real data examples. We compare the performance of StaPLR with an existing view selection method called the group lasso and observe that, in terms of view selection, StaPLR is often more conservative and has a consistently lower false positive rate. 
	
\end{abstract}

\newpage

\section{Introduction}
\label{sect:intro}

Integrating information from different feature sets describing the same set of objects is known as \textit{multi-view learning} \citep{multiview_survey, multiview_review, multiview_bio}. Such different feature sets (\textit{views}) occur naturally in biomedical research as different types of omics data (e.g. genomics, transcriptomics, proteomics, metabolomics) \citep{multiview_bio}, but also as the same profiling data summarized at different levels \citep{costello2014}, or as different gene sets or genetic pathways \citep{wang2010}. In neuroimaging, views may present themselves as different MRI modalities, such as functional MRI and diffusion-weighted MRI \citep{fratello2017}, but also as different feature sets computed from the same structural image \citep{deVos2016}. As there is growing interest in integrating multi-omics and imaging data with other sources of information -- electronic health records, patient databases, and even social media, wearables, and games -- the abundance of multi-view data in biomedical research can only be expected to increase \citep{Fernandez2015, Auffray2016}.
\par
One common problem in biomedical research is a high-dimensional joint classification and feature selection problem, where -- given different classes of objects -- the goal is to identify the features most important for accurate classification \citep{multiview_bio}. When integrating data from multiple views, a typical approach to this problem is \textit{feature concatenation}: simply aggregating the features from all views into one large feature set and fitting a single model to the complete data \citep{multiview_bio}. This is also known as \textit{early integration}, as the views are combined before any further processing \citep{nobel2004, zitnik2019}. Commonly used models for feature selection are generalized linear models (GLMs) with an $L_1$ penalty on the coefficients \citep[\textit{lasso};][]{lasso}, or a mixture of $L_1$ and $L_2$ penalties \citep[\textit{elastic net};][]{elasticnet}. Although these methods can obtain sparse solutions by setting some of the coefficients to zero, they do so without regard to the multi-view structure of the data. This structure is important, as data from a single view is often collected together so that the largest potential savings in time and costs are made by selecting or discarding entire views, rather than individual features. Or, for example, when views correspond to genetic pathways, the most associated gene in a pathway may not necessarily be the best candidate for therapeutic intervention \citep{wang2010}, and selection of complete pathways may be preferable to selecting individual genes. \par
The \textit{group lasso} \citep{grouplasso} is an extension of the lasso which places a penalty on the sum of $L_2$ norms of predefined groups of features, leading to a lasso fit at the view level (i.e. view selection), and shrinkage within views. The group lasso has a single tuning parameter which is typically optimized through cross-validation. It is known, however, that the lasso with prediction-optimal penalty parameter selects too many irrelevant features \citep{Meinshausen2006, Benner2010}. Likewise, it can be observed in the simulation study of \citet{grouplasso} that the group lasso tends to select too many groups. Fitting the group lasso can be slow compared with the regular lasso, as parameter updates are performed block-wise rather than coordinate-wise \citep{statlearn_sparsity, gglasso}. Furthermore, if sparsity within views is desired, an additional mixing parameter needs to be optimized \citep{Simon2013}. \par
The group lasso can be considered a special case of a more general multi-view learning framework known as \textit{multiple kernel learning} \citep{Bach2004, Bach2008}. Multiple kernel learning is one of several approaches to multi-view learning. Other popular approaches include \textit{co-training} style algorithms, which are semi-supervised learning algorithms that use the complementary information in different views to iteratively learn several classifiers that can label observations for each other \citep{cotraining, multiview_book}, and \textit{co-regularization} style algorithms that put a penalty term on the disagreement between classifiers trained on different views, thus forcing consensus among the view-specific classifiers \citep{multiview_review, multiview_book}.
Recently, methods which combine the principles of complementarity and consensus into a single method have been proposed, both in classification (i.e. in multi-view support vector machines \citep{multiview_svm}) and dimension reduction \citep{multiview_cca}. \par 
Another recently popularized multi-view learning framework is that of \textit{multi-view stacking} (MVS) \citep{Li2011, multiview_stacking}. In order to resolve the limitations of the group lasso, we propose an alternative approach to the view selection problem based on this MVS framework. 
MVS is a generalization of \textit{stacking} \citep{Wolpert1992} to multi-view data. In stacking, a pool of learning algorithms (the \textit{base-learners} or first level learners) are fitted to the complete data, and their outputs are combined by another algorithm (the \textit{meta-learner}) to obtain a final prediction. The parameterization of the meta-learner is obtained through training on the cross-validated predictions of the base-learners. Since its inception, stacking has been further studied and expanded upon \citep{Breiman1996, SuperLearner, Subsemble}; a more extensive discussion of stacking is provided by \citet{Sesmero2015}. \par 
In MVS, a base-learner (or pool of base-learners) is trained on each view separately, and a meta-learner is used to combine the predictions of the view-specific models. MVS is thus a \textit{late integration} \citep{nobel2004, zitnik2019} approach to multi-view learning. Several biomedical studies have applied methods which can be considered a form of MVS, showing improved prediction accuracy compared with single-view models and feature concatenation \citep{Li2011, deVos2016, Rahim2016, Liem2017}. Nevertheless, there is no established standard for choosing the base- and meta-learners, and learners which perform well in terms of prediction accuracy often do so at the expense of interpretability. For example, several studies have used random forests as the meta-learner, often with good results in terms of prediction accuracy \citep{Rahim2016,Liem2017,multiview_stacking}, but the resulting models are difficult to interpret and do not allow for easy selection of the most important views. The ability of a researcher or practitioner to understand how or why a classifier makes decisions (i.e. the topic of \textit{explainable AI} \citep{doran2018,holzinger2018}) is important for many real world applications of machine-assisted decision making, particularly in the medical domain \citep{holzinger2018, holzinger2019}. \par
Although applications of MVS have so far focused solely on improving prediction, it also has potential as a group-wise feature selection method. Unfortunately, no unified theoretical underpinning is available regarding the performance of MVS in terms of either prediction accuracy or feature selection. Some progress has been made in terms of the theoretical analysis of the generalization performance of multi-view learning methods \citep{multiview_review}, for example, the derivation of PAC-Bayes bounds for co-regularization style algorithms in a two-view setting \citep{PACbayes}. However, it is not yet clear how these results extend to a setting with more than two views, or whether a similar approach can be used for the theoretical analysis of the multi-view stacking framework. To better understand the MVS approach we introduce \textit{stacked penalized logistic regression} (StaPLR): a special case of MVS where penalized logistic regression is used for both the base-learners and the meta-learner. StaPLR has several advantages over other combinations of base- and meta-learners: logistic regression models are easy to interpret; with appropriately chosen penalties it can be used to perform view selection and/or feature selection within views; and for $L_1$ and $L_2$ penalties the regularization path is fast to compute even for a very large number of features \citep{glmnet}. To perform view selection, StaPLR can be applied with, for example, an $L_2$ penalty at the base level and an $L_1$ penalty at the meta-level, forming a late integration alternative to the group lasso. \par
Of primary interest is whether StaPLR selects the correct views, that is, whether it can separate the views containing signal from those containing only noise. Additionally, it is of interest how the classifiers produced by StaPLR perform in terms of predictive accuracy. The derived results can be used as an indicator of the view selection potential of the general MVS approach. \par
The rest of this article is structured as follows. In Section \ref{sect:MVS} we discuss the multi-view stacking algorithm. In Section \ref{sect:nnc} we verify the importance of nonnegativity constraints on the parameters of the meta-learner for preventing degenerate behavior in MVS with a broad class of base-learners, including penalized GLMs. In Section \ref{sect:StaPLR} we introduce StaPLR as a special case of MVS. In Section \ref{sect:simulations} we compare, on simulated data, the view selection and classification performance of StaPLR with that of the group lasso, and in Section \ref{sect:realdata} we apply both methods to two gene expression data sets. In Section \ref{sect:dicussion} we relate our results on the performance of StaPLR to the general MVS framework, and in Section \ref{sect:conclusions} we present our conclusions. Theoretical proofs are given in the Appendix.

\section{Multi-View Stacking (MVS)} 

\subsection{The MVS Algorithm} \label{sect:MVS}

Let us denote by $\bm{X}^{(1)}, ..., \bm{X}^{(V)}$ a multi-view data set, with $\bm{X}^{(v)}$ the $n \times m_v$ matrix of features in view $v$. Let us denote by $\bm{y} = (y_1, ..., y_n)^T$ the vector of corresponding outcomes. We define a (supervised) learning algorithm or \textit{learner} $A$ as a function that takes as input a labeled data set and produces as output a learned function $\hat{f}$ mapping input vectors to outcomes. \par 
The MVS procedure was already defined and briefly discussed by \citet{Li2011} and \citet{multiview_stacking}. Here, we give a somewhat broader definition of the MVS procedure in Algorithm \ref{al:1}, allowing for general base- and meta-learners, and for multiple learners per view. 
We denote by $A_{v,b}$ the $b$th base-learner for view $v$, with $B_v$ the total number of base-learners for that view. We denote the meta-learner by $A_{\text{meta}}$. Although we consider only a single meta-learner, the procedure could easily be extended by using multiple meta-learners and combining their predictions at even higher levels if desired. \par 
The two key components of training any stacked model are (1) training the base-learners, and (2) training the meta-learner. These can be performed in any order, but in Algorithm \ref{al:1} we first show the training of the base-learners: for each view $\bm{X}^{(v)}$, $v = 1, ..., V$, we apply the base-learners $A_{v,1}, ..., A_{v,B_v}$ to all $n$ observations of that view to obtain a set of learned functions $\hat{f}_{v,1}, ..., \hat{f}_{v,B_v}$. \par 
To train the meta-learner, we need to obtain a set of cross-validated predictions for each learned function $\hat{f}_{v,b}$. Therefore, we partition the data into $K$ groups, and denote by $S_1,S_2,...,S_K$ the $K$-partition of the index set $\{1,2,...,n\}$. For each fold $k = 1, ..., K$, we apply the learner $A_{v,b}$ to the observations which are not in $S_k$, denoted by $\bm{X}^{(v)}_{i \notin S_k}$, $\bm{y}^{}_{i \notin S_k}$. We then apply the learned function $\hat{f}_{v,b,k}$ to the observations in $S_k$ to obtain the corresponding cross-validated predictions. Thus we obtain an $n$-vector of cross-validated predictions for each view and corresponding base-learner, denoted by $\bm{z}^{(v,b)}$. We collect these vectors in an $n \times B$ matrix $\bm{Z}$, where $B = \sum_v B_v$. These cross-validated predictions are then used as the input features for the meta-learner to obtain $\hat{f}_{\text{meta}}$. The final stacked prediction function is then $\hat{f}_{\text{meta}}\big(\hat{f}_{1,1}(\bm{X}^{(1)}), ..., \hat{f}_{V,B_V}(\bm{X}^{(V)})\big)$. \par
It is clear that if the meta-learner is chosen such that it returns sparse models, MVS can be used for view selection. If we choose a single base-learner for each view, the view selection problem is just a feature selection problem involving $V$ features. Compared with feature concatenation, where one has to solve a group-wise feature selection problem involving $\sum_v m_v$ features, this is an easier task. Furthermore, all computations performed on lines \ref{al:1:basetrain} and \ref{al:1:cvtrain} of Algorithm \ref{al:1} are independent across views, base-learners, and cross-validation folds, and can thus be parallelized to improve the scalability of MVS. \par 

\begin{algorithm} 
	\caption{Multi-View Stacking} \label{al:1}
	\DontPrintSemicolon
	\KwData{Views $\bm{X}^{(1)}, \dots, \bm{X}^{(V)}$ and outcomes $\bm{y} = (y_1, \dots, y_n)^T$.}
	\For{$v$ = 1 to $V$}{
		\For{$b$ = 1 to $B_v$}{
			$\hat{f}_{v,b} = A_{v,b}(\bm{X}^{(v)}, \bm{y})$ \label{al:1:basetrain}
		}
	}
	\For{$v$ = 1 to $V$}{
		\For{$b$ = 1 to $B_v$}{
			\For{$k$ = 1 to $K$}{
				$\hat{f}_{v,b,k} = A_{v,b}(\bm{X}^{(v)}_{i \notin S_k}, \bm{y}^{}_{i \notin S_k})$  \label{al:1:cvtrain} \;
				$\bm{z}^{(v,b)}_{i \in S_k} = \hat{f}_{v,b,k}(\bm{X}^{(v)}_{i \in S_k})$
			}
		}
	}
	$\bm{Z} = (\bm{z}^{(1,1)}, \bm{z}^{(1,2)}, \dots, \bm{z}^{(1,B_1)}, \bm{z}^{(2,1)}, \dots, \bm{z}^{(V,B_V)})$\;
	$\hat{f}_{\text{meta}} = A_{\text{meta}}(\bm{Z}, \bm{y})$\;
	$\hat{\bm{y}} = \hat{f}_{\text{meta}} \big(\hat{f}_{1,1}(\bm{X}^{(1)}), \dots, \hat{f}_{1,B_1}(\bm{X}^{(1)}), \hat{f}_{2,1}(\bm{X}^{(2)}), \dots, \hat{f}_{V,B_V}(\bm{X}^{(V)})\big)$\;
\end{algorithm}

\subsection{Nonnegativity Constraints} \label{sect:nnc}

In the context of stacked regression, \citet{Breiman1996} suggested to constrain the parameters of the meta-learner to be nonnegative and sum to one, in order to create a so-called interpolating predictor, i.e. to ensure that the predictions of the meta-learner stay within the range $[\min_b f_b(\bm{x}_i), \max_b f_b(\bm{x}_i)]$ for all observations $\bm{x}_i$, $i = 1, \dots, n$. The sum-to-one constraint proved to be generally unnecessary, but the nonnegativity constraints were crucial in finding the most accurate model combinations \citep{Breiman1996}, a finding corroborated by \citet{Leblanc1996}. However, in a classification context \citet{Ting1999} found that nonnegativity constraints did not substantially affect classification accuracy. \par
Here we provide an additional argument in favor of nonnegativity constraints from a view-selection perspective. Consider MVS with a base-learner for which one of the possible learned functions returns a constant prediction, such as the intercept-only model. Such base-learners include $L_1$- and $L_2$-penalized GLMs. For penalized base-learners, the tuning parameter is often chosen through cross-validation. If we apply a penalized base-learner to some view which contains only noise (i.e. for each feature in this view the true regression coefficient is zero), then it is likely that the model with the lowest cross-validation error is the intercept-only model. \par
Now let us partition a view into $K$ groups, and again denote by $S_1,S_2,...,S_K$ the $K$-partition of the index set $\{1,2,...,n\}$. Assuming that for each fold the fitted model is the linear intercept-only model, the $K$-fold cross-validated predictor $\bm{z}=(z_1,...,z_n)^T$ is given by
\begin{align} \label{eq:z}
z_{i}=\frac{1}{n-|S_k|}\sum_{j\notin S_k} y_j\qquad\text{for all $i\in S_k$, $k=1,...,K$}, 
\end{align}
where $|S_k|$ denotes the cardinality of the set $S_k$. Given the intercept-only model, the cross-validated predictor is not a function of the features in the corresponding view. Therefore, this view should ideally obtain a weight of zero in the meta-learner. However, the cross-validated predictor is not independent of the outcome: in view of Lemma \ref{negcor} the correlation between the cross-validated predictor $\bm{z}$ and the outcome $\bm{y}$ is always negative, with the strength of the correlation increasing with the number of folds.
\begin{lemma} \label{negcor}
	Let $\bm{y}=(y_1,...,y_n)^T$ be the outcome variable, and let $\bm{z}$ be the cross-validated predictor as defined in \eqref{eq:z}. Let $\sigma^2(\bm{y})$ and $\sigma^2(\bm{z})$ be the empirical variance of the vector $\bm{y}$ (i.e. $\sigma^2(\bm{y})=(n-1)^{-1}\sum_{j=1}^n (y_j-\bar{y})^2$, with $\bar{y}=n^{-1}\sum_{j=1}^n y_j$) and the empirical variance of $\bm{z}$ (i.e. $\sigma^2(\bm{z})=(n-1)^{-1}\sum_{j=1}^n (z_j-\bar{z})^2$, with $\bar{z}=n^{-1}\sum_{j=1}^n z_j$), respectively. Then the Pearson correlation between $\bm{y}$ and $\bm{z}$ is equal to
	\[ \rho(\bm{y},\bm{z})=-\frac{\sum_{k=1}^{K}\Big( \sum_{j\in S_k}(y_j-\bar{y})\Big)^2/(n-|S_k|)}{(n-1)\sigma(\bm{y})\sigma(\bm{z})}. \]
	The proof can be found in appendix \ref{sect:proof1}.
\end{lemma}
\begin{corollary}
	In the special case when all folds are of the same size, i.e. $|S_k|=n/K$,
	\[ \rho(\bm{y},\bm{z})=- \frac{(K-1)\sigma(\bm{z})}{\sigma(\bm{y})}. \]
\end{corollary}
\begin{corollary}
	In the special case of leave-one-out cross-validation, i.e. $K = n$,
	\[ \rho(\bm{y},\bm{z})= -1. \]
\end{corollary}
This negative correlation is an artifact of the cross-validation procedure and can produce misleading results in the meta-learner. Consider MVS with two views, $\bm{X}^{(1)}$ and $\bm{X}^{(2)}$, where all features are standard normal, and again a single base-learner for which one of the possible fitted models is the linear intercept-only model.  
Suppose that in truth, the response only depends on the features in $\bm{X}^{(2)}$, e.g. $\bm{y} = \bm{X}^{(2)}\bm{\beta} + \bm{\epsilon}$, with $\bm{\beta}$ a vector of nonzero regression weights, and errors $\bm{\epsilon} = (\epsilon_1,\dots,\epsilon_n)^T$, with $\epsilon_i \overset{\text{iid}}{\sim} \mathcal{N}(0,\sigma^2_{\epsilon})$ for all $i$, and $\sigma^2_{\epsilon} > 0$. Then Lemma \ref{ols_estimates} shows that it can happen that MVS with a linear meta-learner will select the wrong view. 
\begin{lemma} \label{ols_estimates}
	Let $\hat{f}_1$ be the linear intercept-only model, with leave-one-out cross-validated predictor $\bm{z}^{(1)}$, such that $\rho(\bm{y},\bm{z}^{(1)}) = -1$. Let $\hat{f}_2$ be a linear model fitted to $\bm{X}^{(2)}$, with cross-validated predictor $\bm{z}^{(2)}$, such that $0 < \rho(\bm{y},\bm{z}^{(2)}) < 1$. Then for the linear meta-learner $\beta_0 + \beta_{1}\hat{f}_{1}(\bm{X}^{(1)}) + \beta_{2}\hat{f}_{2}(\bm{X}^{(2)})$, the least-squares parameter estimates are
	\[ \hat{\beta}_{1} = 1 - n, \]
	\[ \hat{\beta}_{2} = 0. \]
	The proof can be found in appendix \ref{sect:proof2}.  	
\end{lemma}
In Lemma \ref{ols_estimates} a negative weight is given to $\hat{f}_{1}$ (the intercept-only model), while $\hat{f}_{2}$ (the model containing signal) is excluded from the meta-learner. The selected view is $\bm{X}^{(1)}$, which contains only noise.
Estimating the coefficients using $L_1$- or $L_2$-penalized estimation with tuning parameter selected through cross-validation does not help since the estimated prediction function described in Lemma \ref{ols_estimates} has zero cross-validation error. Cross-validation will therefore always select the least-penalized model under consideration, thus providing no meaningful shrinkage of $\beta_1$. However, nonnegativity constraints can prevent such degenerate behavior by forcing $\beta_1$ to be zero, allowing a nonzero estimate of $\beta_{2}$.   \par
Leave-one-out cross-validation is an extreme case, as for smaller values of $K$ such negative correlations will be lower in magnitude. Furthermore, it should be noted that the cross-validated predictors as described in \eqref{eq:z} occur only in the training phase of the multi-view stacking procedure. As can be observed in Algorithm \ref{al:1}, the matrix of cross-validated predictions $\bm{Z}$ is used to train the meta-learner (line 15), but the final stacked prediction function (line 16) uses the view-specific functions learned from the complete views (lines 1:5). Thus, the final stacked prediction function will not produce the kind of piece-wise constant predictions seen in \eqref{eq:z}. Nevertheless, the introduced correlations can cause the meta-learner to include superfluous views in the model. From a view-selection perspective this is clearly undesirable. \par

\section{Stacked Penalized Logistic Regression (StaPLR)} \label{sect:StaPLR}

\subsection{Penalized Logistic Regression} \label{sect:PLR}

For a binary outcome $\bm{y} = (y_1, y_2, \dots, y_n)^T \in \{0,1\}^n$, and a feature set $\bm{X} = (x_{ij}) \in \mathbb{R}^{n \times m}$, the logistic regression model is given by
\begin{equation}
\text{Pr}(y_i = 1 | \bm{x}_i) = \frac{1}{1 + \text{exp}(-\beta_0 - \bm{\beta}^T \bm{x}_i)},
\end{equation}
where $\bm{x}_i = (x_{i1}, x_{i2}, \dots, x_{im})^T$ is the feature vector corresponding to observation $i$, $\beta_0 \in \mathbb{R}$, and $\bm{\beta} \in \mathbb{R}^m$. Parameter estimates are typically obtained through maximum likelihood estimation. In penalized estimation, a penalty term on $\bm{\beta}$ is applied in the optimization problem. We write the intercept $\beta_0$ separately, as it is usually not penalized. For example, in the case of an $L_2$ penalty the parameters are estimated as
\begin{equation} \label{eq:plr_base_ridge}
\hat{\beta}_0, \hat{\bm{\beta}} = \argmax_{\beta_0, \bm{\beta}} \left\{ \sum_{i = 1}^{n} \left[ y_i(\beta_0 + \bm{\beta}^T \bm{x}_i) - \log(1 + \text{exp}(\beta_0 + \bm{\beta}^T \bm{x}_i)) \right] - \lambda\|\bm{\beta}\|_2^2 \right\},
\end{equation}
where $\lambda \geq 0$ is a tuning parameter. In the case of an $L_1$ penalty the rightmost term is replaced by $\lambda\|\bm{\beta}\|_1$. \par
A suitable value of $\lambda$ is generally chosen through cross-validation. First, one defines a set of, say, 100 candidate values of $\lambda$. Next, for each value of $\lambda$, cross-validation is applied to obtain an estimate of the associated out-of-sample error, and the value of $\lambda$ with lowest cross-validation error is selected. Finally, \eqref{eq:plr_base_ridge} is optimized using the complete data and the selected value of $\lambda$ to obtain the final model. It should be noted that, since the value of $\lambda$ is chosen such that it minimizes cross-validation error, the associated error estimate most likely underestimates the true out-of-sample error \citep{varma2006}. Using cross-validation to choose the tuning parameter should thus be considered part of the learning process. In order to validate such a model using cross-validation, a double cross-validation is required, where an inner loop used to select the tuning parameter is nested in an outer validation loop \citep{varma2006}.

\begin{algorithm} 
	\caption{Pseudocode for a penalized logistic regression learner} \label{al:2}
	\DontPrintSemicolon
	\KwIn{A set of features $\bm{X}$, and binary outcomes $\bm{y}$.}
	\KwOut{A learned function $\hat{f}$.}
	Define a set of candidate tuning parameter values $\bm{\Lambda}$. \;
	Randomly split the data into $K$ cross-validation folds of roughly equal size. \;
	\ForEach{$\lambda \in \bm{\Lambda}$}{
		\For{$k = 1$ to $K$}{
			Optimize the penalized likelihood (e.g. equation \eqref{eq:plr_base_ridge}) using only the observations outside of fold $k$. \;
			Apply the trained model to obtain an error for each observation in fold $k$.
		}
		Average the cross-validation error across all observations.
	}
	Choose $\lambda^*$ to be the value in $\bm{\Lambda}$ with lowest cross-validation error. \;
	Optimize the penalized likelihood using all observations and penalty parameter value $\lambda^*$ to obtain a set of parameters $\hat{\beta}_0^{(\lambda^*)}$, $\hat{\bm{\beta}}^{(\lambda^*)}$. \;
	Return the learned function $\hat{f}(\bm{X}) = 1 / (1 + \exp(- \hat{\beta}_0^{(\lambda^*)} - \bm{X}\hat{\bm{\beta}}^{(\lambda^*)}))$.
\end{algorithm}

\subsection{Stacked Penalized Logistic Regression}

We previously defined a (supervised) learner $A$ as a function that takes as input a labeled data set and produces as output a learned function $\hat{f}$ mapping input vectors to outcomes. Thus the procedure of fitting a penalized logistic regression model (including cross-validation for $\lambda$) described in section \ref{sect:PLR} can be considered a learner. Such a learner is described in pseudocode in Algorithm \ref{al:2}. We define StaPLR as a special case of MVS where all learners are penalized logistic regression learners. StaPLR thus denotes a special case of Algorithm \ref{al:1}, where all $A_{v,b}$ (lines 3 and 9) and $A_{\text{meta}}$ (line 15) are functions of the form described in Algorithm \ref{al:2}. Note that in Algorithm \ref{al:2}, $\bm{X}$ refers to the first input argument of the learner. When applied as a base-learner in a stacked model, this would be a view $\bm{X}^{(v)}$, and when applied as the meta-learner, it would be the matrix of cross-validated predictions $\bm{Z}$.
In the remainder of this article we use a single base-learner with an $L_2$ penalty which we apply to every view. For the meta-learner we choose an $L_1$ penalty. This way we induce sparsity at the view level and shrinkage within each view, thus providing the configuration most similar to the group lasso model. We use probabilities rather than hard classifications as input for the meta-learner. These values contain information about the uncertainty of the predictions and were previously found to work better in stacked generalization than hard class labels \citep{Ting1999}. In order to preserve this information, and because the predictions of the base-learners are already on a common scale, we do not standardize the inputs to the meta-learner. Parameters are estimated using coordinate descent \citep{glmnet}. For both the base and meta-learner's internal cross-validation loops we use $K = 10$. The set of candidate tuning parameter values $\Lambda$ is a sequence of 100 values adaptively chosen by the software \citep{glmnet}. \par 
We demonstrate in our simulations that the addition of nonnegativity constraints on the parameters of the meta-learner improves the view selection performance of StaPLR. When differentiating between StaPLR with and without nonnegativity constraints we use the notation $\text{StaPLR}^+$ and $\text{StaPLR}^-$, respectively. In coordinate descent, nonnegativity constraints are easily implemented by simply setting coefficients to zero if they become negative during the update cycle \citep{glmnet,statlearn_sparsity}.

\FloatBarrier

\color{black}
\section{Simulations} \label{sect:simulations}

In this section we compare, on simulated data, the performance of StaPLR with that of the group lasso. In Subsection \ref{sect:viewselection} we investigate the view selection performance of both methods under a number of experimental conditions, and in Subsection \ref{sect:classification} we evaluate the obtained classifiers in terms of area under the receiver operating characteristic curve (AUC). In Subsection \ref{sect:largersamplesizes} we investigate the view selection performance of both methods for larger sample sizes, and in Subsection \ref{sect:viewsizes} we explore how the number of features in a view affects the view selection performance if the amount of signal strength is kept constant. \par 
All simulations are performed in R (version 3.4.0) \citep{R}. Penalized logistic regression models are fitted using the package \texttt{glmnet} 1.9-8 \citep{glmnet}. The (logistic) group lasso is fitted using the package \texttt{gglasso} 1.3 \citep{gglasso}. 

\subsection{View Selection Performance} \label{sect:viewselection}

We investigate the ability of StaPLR and the group lasso to select the correct views. We use two different sample sizes ($n$ = 200 or 2000) and two different view sizes ($m_v$ = 250 or 2500). We use block correlation structures defined by two parameters, namely the population correlation between features in the same view $\rho_w$, and the population correlation between features in different views $\rho_b$. We use three different parameterizations: ($\rho_w = 0.1, \rho_b = 0$), ($\rho_w = 0.4, \rho_b = 0$), and ($\rho_w = 0.4, \rho_b = 0.2$), for a total of $2 \times 2 \times 3 = 12$ experimental conditions. \par 
We generate 30 disjoint views of equal size $\bm{X}^{(v)}$, $v = 1 \dots 30$, with each view an $n \times m_v$ matrix consisting of normally distributed features scaled to zero mean and unit variance.
Within each view, we randomly determine which features correspond to signal (i.e. have a true relation with the response) and which correspond to noise. In 5 views, the probability that a feature corresponds to signal is 1. In another 5 views, the probability that a feature corresponds to signal is 0.5. In the remaining 20 views, the probability that a feature corresponds to signal is 0. 
Denote by $m = \sum_v m_v$ the total number of features. For each feature $\bm{x}_j = (x_{1j}, x_{2j}, \dots, x_{nj})^T$, $j = 1 \dots m$, we then determine a regression weight $\theta_j$. If $\bm{x}_j$ corresponds to signal, $\theta_j$ = 0.04 or -0.04, each with probability 0.5. This effect size was chosen because simulations showed that with $n = 200$, $m_v = 250$, and ($\rho_w = 0.4, \rho_b = 0$), the class probability distribution is approximately uniform. If $\bm{x}_j$ corresponds to noise, $\theta_j = 0$. We then determine class probabilities $p_i = 1/(1 + \exp(-\bm{\theta}^T\bm{x}_i))$, with $\bm{\theta} = (\theta_1, ... \theta_m)^T$, and class labels $y_i \sim \text{Bernoulli}(p_i)$. \par
\begin{sidewaysfigure}
	\begin{subfigure}{0.49\textheight}
		\centering
		\includegraphics{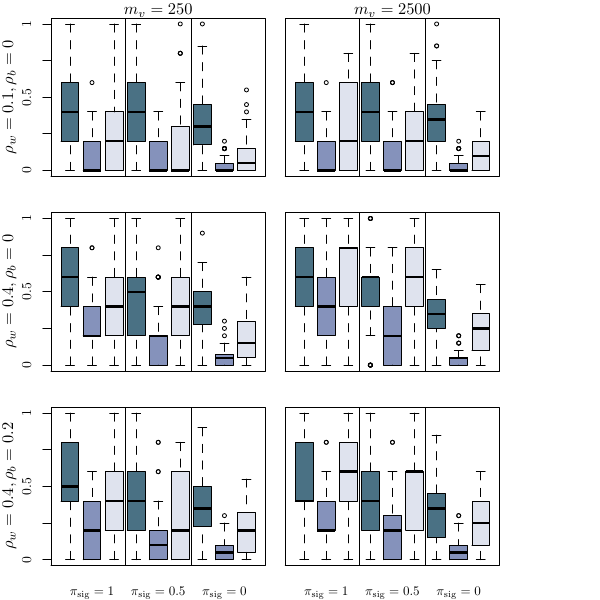}
		\caption{$n = 200$	\label{fig:e1n200}}
	\end{subfigure}
	\begin{subfigure}{0.49\textheight}
		\centering
		\includegraphics{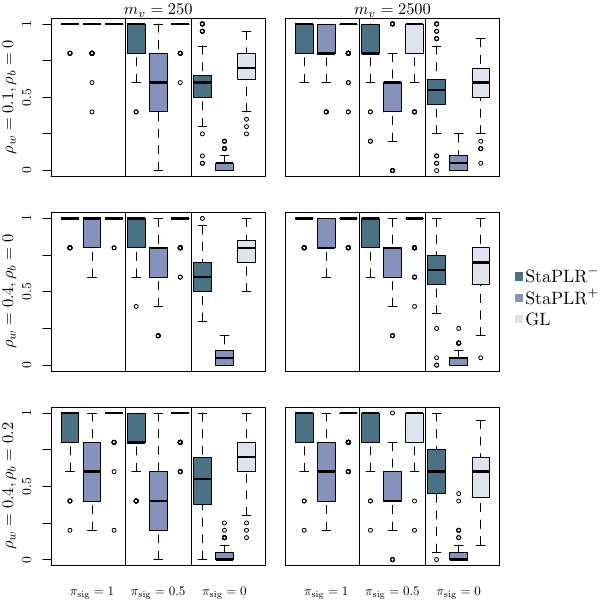}
		\caption{$n = 2000$ \label{fig:e1n2000}}	
	\end{subfigure}
	\caption{Box plots of the observed inclusion probabilities for views with different proportions of signal (denoted by $\pi_{sig}$). The within-view correlation is denoted by $\rho_w$, the between-view correlation by $\rho_b$, and the number of features per view by $m_v$.  \label{fig:e1}}	
\end{sidewaysfigure}

The aim of applying StaPLR and the group lasso is to select the views which contain signal, and discard the others. We calculate the observed probability of a view with a certain proportion of signal (0, 0.5, or 1) being included in the final model. We perform 100 replications per condition. Box plots over all replications are shown in Figure \ref{fig:e1}. It can be observed that StaPLR with nonnegativity constraints ($\text{StaPLR}^+$) maintains a lower false positive rate than the group lasso regardless of sample size, number of features or correlation structure, with the largest differences seen in the $n = 2000$ case (Figure \ref{fig:e1n2000}). It is, however, also more conservative, selecting fewer views containing signal. Without the nonnegativity constraints, $\text{StaPLR}^-$ sometimes has a higher false positive rate than the group lasso, particularly when $n$ is small (Figure \ref{fig:e1n200}). \par

\subsection{Classification Performance} \label{sect:classification}

\begin{sidewaysfigure}
	\begin{subfigure}{0.49\textheight}
		\centering
		\includegraphics{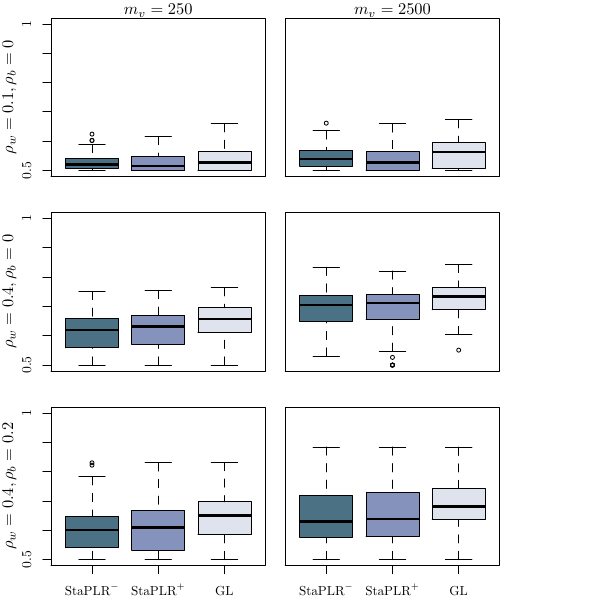}
		\caption{$n = 200$	\label{fig:e4n200}}
	\end{subfigure}
	\begin{subfigure}{0.49\textheight}
		\centering
		\includegraphics{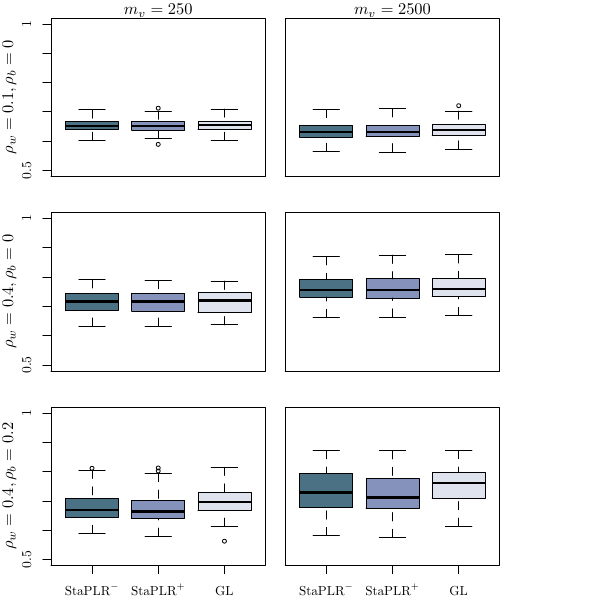}
		\caption{$n = 2000$ \label{fig:e4n2000}}	
	\end{subfigure}
	\caption{Box plots of the AUC values for 100 test sets per condition. The within-view correlation is denoted by $\rho_w$, the between-view correlation by $\rho_b$, and the number of features per view by $m_v$. \label{fig:e4}}	
\end{sidewaysfigure}

For each replication of each of the 12 conditions from the previous experiment, we generate a test set of size $n = 1000$. The test set is only used for model evaluation; all model fitting including the selection of tuning parameters is performed using (partitions of) the training set. We calculate the AUC on the test set for each of the three methods. It can be observed in Figure \ref{fig:e4} that the different methods have a comparable performance when $n = 2000$ and the features from different views are not correlated. When $n = 200$, or when the features from different views are correlated, the group lasso obtains a slightly higher median AUC. 

\subsection{Larger Sample Sizes} \label{sect:largersamplesizes}

\begin{figure}[t!]
	\centering
	\includegraphics{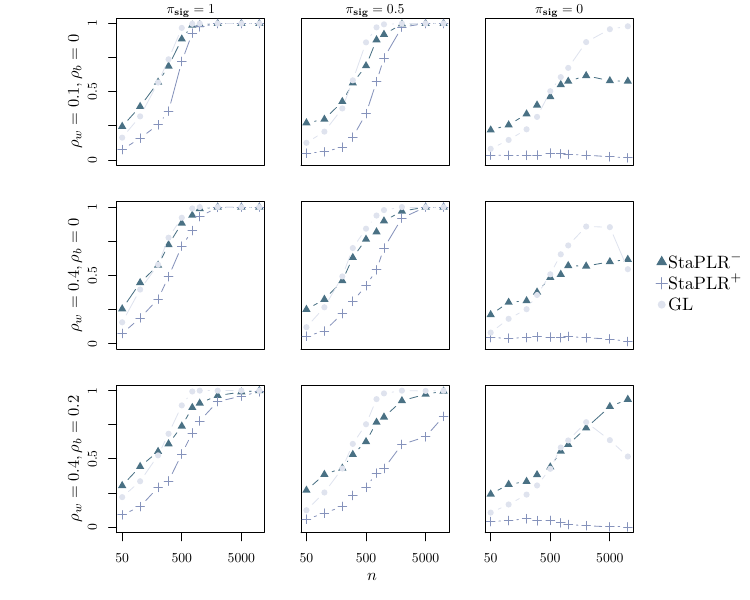}
	\caption{Average inclusion probabilities for views with different proportions of signal (denoted by$\pi_{sig}$), separated by method and correlation structure, across a range of sample sizes. The different sampling points are $n =$ 50, 100, 200, 300, 500, 750, 1000, 2000, 5000 and 10000. Note that the distances along the x-axis are on a $\log_{10}$ scale. \label{fig:e2}}
\end{figure}

In this experiment we investigate the view selection behavior of StaPLR and the group lasso for larger sample sizes $n$. We again use 30 views, but now with 25 features per view, and regression weights $\theta_j$ = 0.12 or -0.12 if $\bm{x}_j$ corresponds to signal, to create a smaller problem which is more easily upscaled to larger sample sizes. We consider ten different sample sizes ranging between 50 and 10000, and again calculate the average inclusion probabilities for views with a certain proportion of signal (0, 0.5, or 1). It can be observed in Figure \ref{fig:e2} that as $n$ increases, $\text{StaPLR}^+$ has an increased probability of selecting views containing signal, while the probability of selecting views containing only noise remains low and even decreases slightly. In contrast, the group lasso and $\text{StaPLR}^-$ have an increased probability of selecting both signal and noise views as $n$ increases. Only in some cases for very high values of $n$ is a decrease in the false positive rate observed. $\text{StaPLR}^+$ consistently has the lowest false positive rate, although for lower sample sizes it is also generally more conservative, selecting fewer views containing signal. However, at high values of $n$, $\text{StaPLR}^+$ often perfectly distinguishes signal and noise.

\subsection{Different View Sizes} \label{sect:viewsizes}

In this experiment we investigate the view selection behavior of StaPLR and the group lasso when views of different sizes are considered at the same time. We consider five different view sizes: 10, 50, 250, 750 and 2500 features. For each view size, we generate one view with signal proportion 1, one view with signal proportion 0.5, and 4 views with signal proportion 0, for a total of 21,360 features across 30 views. We use sample size $n = 2000$. Denote by $\bm{x}^{(v)}_j$, $j = 1, \dots, m_v$, the $j$th feature in view $v$. If $\bm{x}^{(v)}_j$ corresponds to signal, its regression weight $\theta^{(v)}_j$ = $1 / \sqrt{m_v}$ or $-1 / \sqrt{m_v}$, each with probability 0.5. The rest of the coefficients are set to zero. \par
\begin{figure}[t!]
	\centering
	\includegraphics{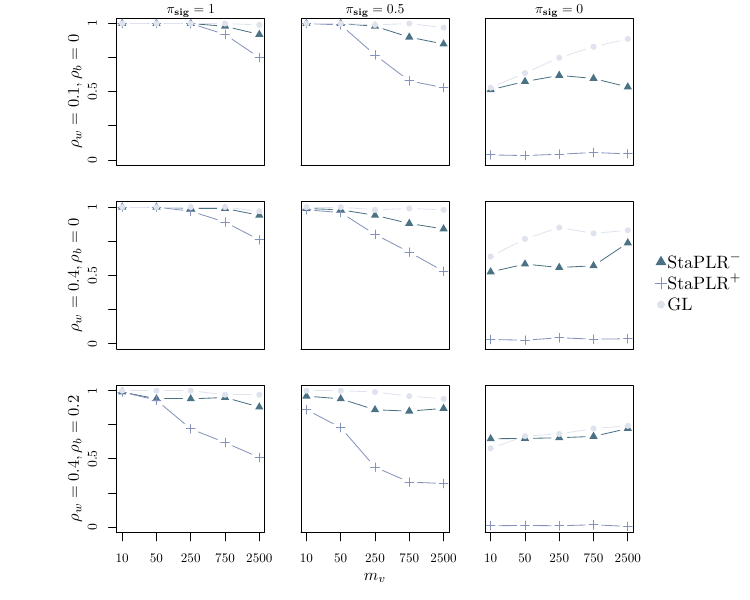}
	\caption{Average inclusion probabilities for views with different proportions of signal (denoted by $\pi_{sig}$), separated by method and correlation structure, as a function of view size. \label{fig:e3}}
\end{figure}

 The results can be observed in Figure \ref{fig:e3}. Again, $\text{StaPLR}^+$ has the lowest false positive rate, and the inclusion probability of a view containing only noise does not appear to depend on its size. In contrast, the group lasso appears to select large views containing only noise more often than smaller views containing only noise. For views containing signal, it can be observed that larger views are less likely to be included by $\text{StaPLR}^+$. This indicates that for views with the same amount of signal strength in an $L_2$ sense, $\text{StaPLR}^+$ favors views containing less features.                                                                                                                                                                                             

\FloatBarrier

\section{Application to Gene Expression Data} \label{sect:realdata}

One type of multi-view data occurs in gene expression profiling where genes can be divided into gene sets based on, for example, signaling pathway involvement or cytogenetic position \citep{C1}. We base our experiments on the real data examples of \citet{Simon2013} by applying StaPLR and the group lasso to two gene expression data sets: the colitis data of \citet{colitis}, and the breast cancer data of \citet{breastcancer}. \par
\begin{figure}[t!]
	\centering
	\includegraphics{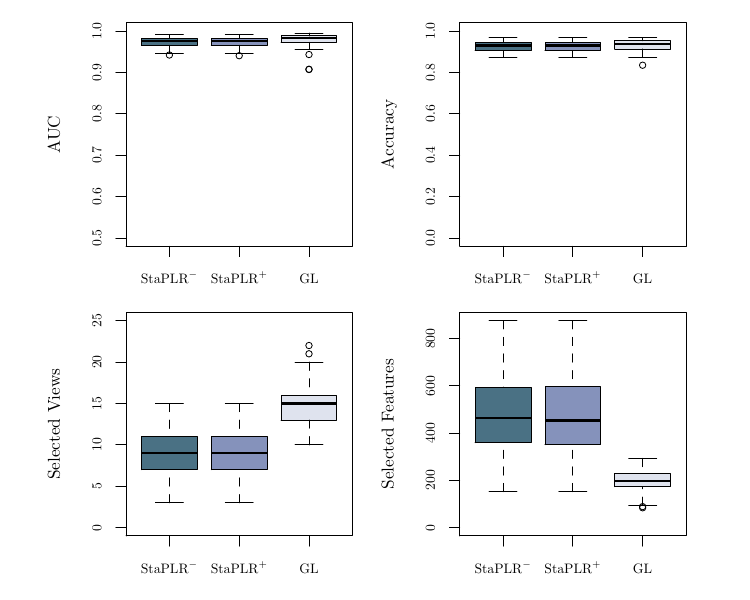}
	\caption{Results of applying StaPLR and group lasso to the colitis gene expression data, in terms of AUC, accuracy, number of selected views and number of selected features.  \label{fig:colitis}}
\end{figure}

The colitis data \citep{colitis} consists of 127 patients: 85 colitis cases (ulcerative colitis or Crohn's disease) and 42 healthy controls. For each patient, gene expression data was collected using an Affymetrix HG-U133A microarray, containing 22,283 probe sets. We matched this data to the C1 cytogenetic gene sets as available from MSigDB 6.1 \citep{C1}. We removed any duplicate probes, any genes not included in the C1 gene sets, and any gene sets for which only a single gene was found in the colitis data. Our final feature matrix consisted of 11,761 genes divided across 356 gene sets, with an average of 33 genes per set. All expression levels were $\log_2$-transformed, then standardized to zero mean and unit variance. In \citet{Simon2013}, the data was randomly split into a training and test set. We apply a similar strategy, randomly splitting the data into two parts of roughly equal size, then using a model fitted to one part to predict the other (i.e. 2-fold cross-validation). The model fitting includes the selection of all tuning parameters through internal cross-validation loops, and the left-out data is used only for model evaluation. Additionally, we repeat this process 50 times to account for variability due to the random partitioning. We thus obtain 50 sets of predictions for each of the three methods:  $\text{StaPLR}^-$, $\text{StaPLR}^+$, and the group lasso. We calculate both classification accuracy (using a cut-off of .5) and AUC. Additionally, for each of the $50 \times 2$ fitted models, we record the number of selected views (gene sets) and features (genes). The results can be observed in Figure \ref{fig:colitis}. All methods have comparable performance in terms of AUC and accuracy, although the group lasso obtains slightly higher median scores than StaPLR. However, StaPLR selects fewer views but more features, whereas the group lasso selects more views but fewer features. The differences between $\text{StaPLR}^+$ and $\text{StaPLR}^-$ appear negligible.

\begin{figure}[t!]
	\centering
	\includegraphics{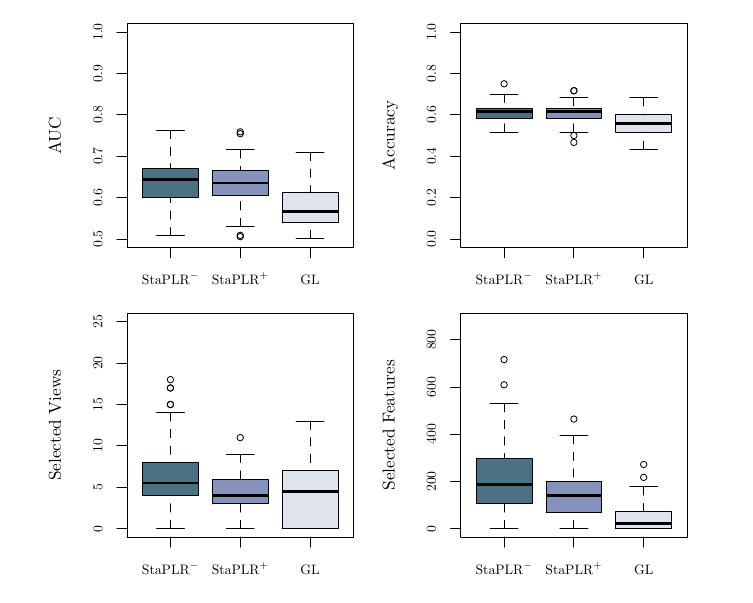}
	\caption{Results of applying StaPLR and group lasso to the breast cancer gene expression data, in terms of AUC, accuracy, number of selected views and number of selected features.  \label{fig:breastcancer}}
\end{figure}

The breast cancer data \citep{breastcancer} consists of 60 tumor samples of patients diagnosed with estrogen positive breast cancer treated with tamoxifen for 5 years, and are labeled according to whether the patients were disease free (32 cases) or cancer recurred (28 cases). For each sample, gene expression data was collected using an Arcturus 22k microarray. We applied the same procedure of matching the gene expression data to the C1 gene sets, obtaining a feature matrix of 12,722 genes divided across 354 sets, with an average of 36 genes per set. As the data was already $\log_2$-tranformed, we only standardized each feature to zero mean and unit variance. The results can be observed in Figure \ref{fig:breastcancer}. StaPLR, both with and without nonnegativity constraints, outperforms the group lasso in terms of AUC and accuracy. The differences between $\text{StaPLR}^+$ and $\text{StaPLR}^-$ in terms of AUC and accuracy are negligible, but the addition of nonnegativity constraints leads to fewer views and fewer features selected on average. Compared with the group lasso, $\text{StaPLR}^+$ selects on average a similar number of views, but a larger number of features. However, this is in part caused by the fact that the group lasso selects no views at all in 27\% of the produced models, whereas $\text{StaPLR}^+$ selects no views in only 4\% of the models.

\FloatBarrier

\section{Discussion} \label{sect:dicussion}

In our real data examples, $\text{StaPLR}^+$ provided similar or better classification accuracy than the group lasso, while selecting a similar or lower number of views. Although the models produced by $\text{StaPLR}^+$ in the colitis data were sparser at the view level than those produced by the group lasso, they were not sparser at the feature level: $\text{StaPLR}^+$ tended to select a smaller number of views with a larger number of features, while the group lasso tended to select a larger number of views with a smaller number of features. In our simulation experiments on different view sizes (Section \ref{sect:viewsizes}) we observed that $\text{StaPLR}^+$ favored views containing fewer features, under the condition that the views contained the same amount of signal strength in an $L_2$ sense. Although this condition allowed us to investigate the effect of the number of features in isolation from the effect of signal strength, it is unlikely for such a condition to be satisfied in real data. In its presented form, StaPLR does not explicitly favor smaller views, but if such behavior is desired a scaling factor depending on the number of features can easily be added. \par
In this article we chose a specific parameterization of StaPLR aimed at selecting or discarding entire views. If sparsity within views is desired this could be achieved by, for example, employing an $L_1$ penalty for the base-learner. This indicates that StaPLR may also form an alternative to other complex penalties such as the sparse group lasso \citep{Simon2013}. \par
\begin{figure}[h!]
	\centering
	\includegraphics{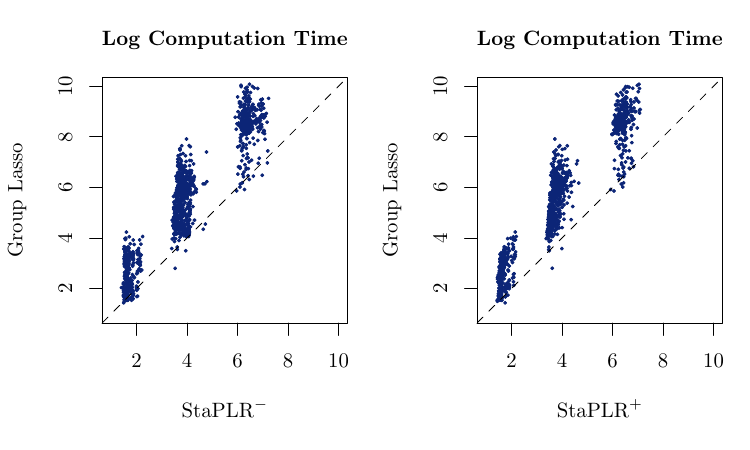}
	\caption{Comparison of the log computation time (in minutes) of $\text{StaPLR}^+$ and $\text{StaPLR}^-$ with the group lasso. Each point represents one replication of the experiments described in Section \ref{sect:viewselection}. The different clouds occur due to the different experimental conditions. Note that the base learners of the StaPLR models were fitted sequentially rather than in parallel. In nearly all cases $\text{StaPLR}^+$ and $\text{StaPLR}^-$ were faster than the group lasso, often considerably so. \label{fig:computation_time}}
\end{figure}

 
In our simulations, fitting models using StaPLR was considerably faster than the group lasso although in practice this will depend on the number of views, view size, and available computational resources. As all simulations were performed on a batch scheduling cluster, and such an environment is not ideal for comparing computation speed, we opted not to include a formal speed comparison in our results. However, a plot comparing the computation time of both methods can be observed in Figure \ref{fig:computation_time}. Our experience suggests that in the event of a small number of views compared with the number of features per view, a large speed-up can be gained even without any parallelization, by using coordinate-wise rather than block-wise updating. Parallelization can increase this computational speed advantage even further. \par

The parameter space of StaPLR is very restricted compared with applications of MVS with multiple or more complex base-learners. We chose logistic regression as the base learner since it is among the best-known classifiers in a variety of scientific fields, and because it produces class probabilities rather than simple class labels to use as input for the meta-learner. In a real life setting, one could choose a different base-learner for each view based on domain knowledge, or even try to learn the best base-learner for each view from a set of candidates. More complex base-learners may be able to capture non-linear relationships and increase predictive performance, but this often comes at a cost in terms of model interpretability. \par
We chose the lasso to perform view selection since it is fast to train and remains a very popular method. Despite its known drawbacks, we found in our experiments that when the lasso was used as the meta-learner in an MVS model with additional nonnegativity constraints, the probability of including superfluous views remained low across all experimental conditions. However, it was generally more conservative than the group lasso, also selecting fewer views containing signal. The performance of the method could potentially be increased by changing the meta-learner. For example, by using a two-step procedure such as the \textit{adaptive lasso} \citep{adaptive_lasso}, a nonconvex penalty such as SCAD \citep{SCAD}, or a subsampling method such as \textit{stability selection} \citep{stability_selection}. One could even apply multiple feature-selecting meta-learners and combine their results using some aggregation method \citep{bolon2019}. These alternatives can be considerably more computationally expensive than the regular lasso model. However, the use of such methods is still more feasible when applied to the MVS meta-learning problem (which only has a number of features equal to the number of views) than when applied to the full group-wise feature selection problem. \par 
In this article we applied view selection using an MVS-based method in the context of logistic regression. In the future, it would be interesting to investigate how view selection can be applied in the context of other classifiers such as multi-view support vector machines \citep{multiview_svm}, or in the context of multi-view dimension reduction \citep{multiview_cca}.

\section{Conclusions} \label{sect:conclusions}

We have introduced stacked penalized logistic regression (StaPLR) as a late integration view selection method based on multi-view stacking. We have further motivated the use of nonnegativity constraints on the parameters of the meta-learner in multi-view stacking with a broad class of base-learners, and shown that such constraints improve the view selection performance of StaPLR. Compared with the group lasso, our simulations have shown that StaPLR with nonnegativity constraints produces sparser models with a comparable, though in some cases slightly reduced, classification performance. This slight reduction in classification performance may be caused by the fact that StaPLR with nonnegativity constraints is often more conservative, also selecting fewer views containing signal than the group lasso. However, it has a much lower false positive rate in terms of the selected views. In our real data examples, StaPLR provided similar or better classification accuracy than the group lasso, while selecting a similar or lower number of views. Our results indicate that multi-view stacking can be used not only to improve model accuracy, but also to select those views that are most important for prediction. By combining view selection with interpretable classifiers we come closer to truly explainable methods for multi-view learning.

\appendix
\FloatBarrier

\section{Proof of Lemma 1, Corollary 1.1 and 1.2} \label{sect:proof1}
The Pearson correlation between the cross-validated predictor $\bm{z}$ and the outcome $\bm{y}$ is defined as
\begin{align*}
\rho(\bm{y},\bm{z})&=\frac{\sum_{j=1}^{n}(y_j-\bar{y})(z_j-\bar{z})}{(n-1)\sigma(\bm{y})\sigma(\bm{z})}.
\end{align*}
To show that this correlation is always negative we introduce a change of variables. Let $a_j=y_j-\bar{y}$, $j=1,...,n$, and $\bm{b}=(b_1,b_2,...,b_n)$ be the corresponding $K$-fold cross-validated predictor. Let us denote by $b_k^*$ the value of the predictor in group $S_k$, $k = 1,...,K$. Note that $b_j=z_j-\bar{y}$ and that $\sigma(\bm{z})=\sigma(\bm{b})$, $\sigma(\bm{y})=\sigma(\bm{a})$. Then
\begin{align*}
\rho(\bm{y},\bm{z})=\rho(\bm{a},\bm{b})= \frac{\sum_{j=1}^{n}a_j(b_j-\bar{b})}{(n-1)\sigma(\bm{a})\sigma(\bm{b})}
=\frac{\sum_{k=1}^{K} (b_k^*-\bar{b})\sum_{j\in S_k}a_j}{(n-1)\sigma(\bm{a})\sigma(\bm{b})}.
\end{align*}
By noting that $\sum_{k=1}^K \sum_{j\in S_k}a_j=\sum_{j=1}^n a_j= \sum_{j=1}^n y_j-n\bar{y}=0$ and $\sum_{j\in S_k}a_j=-\sum_{j\notin S_k} a_j$ we get that the numerator on the right hand side of the preceding display is further equal to
\begin{align*}
\begin{split}
\sum_{k=1}^{K} (b_k^*-\bar{b})\sum_{j\in S_k}a_j&= \sum_{k=1}^{K}\Big( b_k^*\sum_{j\in S_k}a_j\Big)\\
&= \sum_{k=1}^{K}\Big( \frac{\sum_{j\notin S_k}a_j}{n-|S_k|} \sum_{j\in S_k}a_j\Big)\\
&= -\sum_{k=1}^{K}\frac{(\sum_{j\in S_k}a_j)^2}{n-|S_k|} ,
\end{split}
\end{align*}
where the term on the right hand side is smaller than or equal to zero and it is exactly zero if in all folds $S_k$, $k=1,...,K$, the sum of $a_j$ is zero (i.e. $\sum_{j\in S_k}a_j=0$). This means that in all folds the average of the observations has to be the same as the total average $\bar{y}$, otherwise the correlation between the vectors will be negative. The final formula for the correlation is then given by
\begin{align*}
\rho(\bm{y},\bm{z})=-\frac{\sum_{k=1}^{K}\Big( \sum_{j\in S_k}(y_j-\bar{y})\Big)^2/(n-|S_k|)}{(n-1)\sigma(\bm{y})\sigma(\bm{z})}.
\end{align*}
Next we consider the special case when all folds are of the same size (i.e. $|S_k|=n/K$, for all $k=1,...,K$). Note that in this case $\bar{z}=\bar{y}$, and the formula simplifies to
\begin{align*}
\begin{split}
\rho(\bm{y},\bm{z})&=-\frac{\sum_{k=1}^{K}\Big( n\bar{y}-\sum_{j\notin S_k}y_j -|S_k|\bar{y}\Big)^2/(n-|S_k|)}{(n-1)\sigma(\bm{y})\sigma(\bm{z})}\\
&=-\frac{\sum_{k=1}^{K}(n-|S_k|) \Big( \bar{z}-z_k^*\Big)^2}{(n-1)\sigma(\bm{y})\sigma(\bm{z})}\\
&=- \frac{(K-1)\sum_{k=1}^{K} |S_k| \Big( \bar{z}-z_k^*\Big)^2}{(n-1)\sigma(\bm{y})\sigma(\bm{z})}\\
&=- \frac{(K-1)\sigma(\bm{z})}{\sigma(\bm{y})}.
\end{split}
\end{align*} 
In the special case of leave-one-out cross-validation $K = n$. To show that in this case $\rho(\bm{y},\bm{z}) = -1$ it suffices, by the preceding display, to show that $\sigma(\bm{z}) = \sigma(\bm{y})/(n-1)$:
\begin{align*}
\sigma^2(\bm{z}) &= \frac{1}{n -1} \sum_{j=1}^n \Big( z_{j} - \bar{z} \Big)^2 \\
&= \frac{1}{n -1} \sum_{j=1}^n \Big( \bar{y}_{(-j)} - \bar{y} \Big)^2 \\
&= \frac{1}{n -1} \sum_{j=1}^n \Big( \frac{n\bar{y}}{n-1} - \frac{y_j}{n-1} - \bar{y} \Big)^2 \\
&= \frac{1}{(n -1)^3} \sum_{j=1}^n \Big( \bar{y} - y_j \Big)^2 \\
&= \frac{\sigma^2(\bm{y})}{(n-1)^2}.
\end{align*}

\section{Proof of Lemma 2} \label{sect:proof2}

Let $\hat{f}_1$ be the linear intercept-only model, with leave-one-out cross-validated predictor $\bm{z}^{(1)}$, such that $\rho(\bm{y},\bm{z}^{(1)}) = -1$. Let $\hat{f}_2$ be a second model, with leave-one-out cross-validated predictor $\bm{z}^{(2)}$, where $0 < \rho(\bm{y},\bm{z}^{(2)}) < 1$. Since $\rho(\bm{y},\bm{z}^{(1)}) = -1$, it follows that $\rho(\bm{z}^{(1)},\bm{z}^{(2)}) = -  \rho(\bm{y},\bm{z}^{(2)})$. \par 
For the linear meta-learner $\beta_0 + \beta_{1}\hat{f}_1(\bm{X}^{(1)}) + \beta_{2}\hat{f}_2(\bm{X}^{(2)})$, the least-squares parameter estimate of $\beta_1$ is given by \citep{fox2008}
\begin{align*}
\hat{\beta}_1 = \frac{1}{\gamma} \Big( \sigma^2(\bm{z}^{(2)}) \text{cov}(\bm{y}, \bm{z}^{(1)}) - \text{cov}(\bm{z}^{(1)}, \bm{z}^{(2)})\text{cov}(\bm{y}, \bm{z}^{(2)}) \Big),
\end{align*}
where $\sigma^2(\cdot)$ denotes the empirical variance, $\text{cov}(\cdot,\cdot)$ denotes the empirical covariance, and $\gamma = (1 - \rho^2(\bm{z}^{(1)}, \bm{z}^{(2)})) \sigma^2(\bm{z}^{(1)}) \sigma^2(\bm{z}^{(2)}) $. We can rewrite this in terms of correlations as
\begin{align*}
\begin{split}
\hat{\beta}_1 
&= \frac{1}{\gamma} \Big( \rho(\bm{y},\bm{z}^{(1)}) - \rho(\bm{z}^{(1)}, \bm{z}^{(2)})\rho(\bm{y},\bm{z}^{(2)}) \Big) \sigma^2(\bm{z}^{(2)})\sigma(\bm{z}^{(1)})\sigma(\bm{y})\\
&= \frac{( \rho^2(\bm{z}^{(1)}, \bm{z}^{(2)}) - 1) \sigma^2(\bm{z}^{(2)})\sigma(\bm{z}^{(1)})\sigma(\bm{y})}{(1 - \rho^2(\bm{z}^{(1)}, \bm{z}^{(2)}) ) \sigma^2(\bm{z}^{(2)})\sigma^2(\bm{z}^{(1)})}\\
&= -\frac{\sigma(\bm{y})}{\sigma(\bm{z}^{(1)})}\\
&= -\frac{\sigma(\bm{y})}{\sigma(\bm{y}) / (n - 1)}\\
&= 1 - n.
\end{split}
\end{align*}
Analogously, we can obtain the least-squares estimate of $\beta_2$:
\begin{align*}
\begin{split}
\hat{\beta}_2 
&= \frac{1}{\gamma} \Big( \rho(\bm{y},\bm{z}^{(2)}) - \rho(\bm{z}^{(1)}, \bm{z}^{(2)})\rho(\bm{y},\bm{z}^{(1)}) \Big) \sigma^2(\bm{z}^{(1)})\sigma(\bm{z}^{(2)})\sigma(\bm{y})\\
&= \frac{1}{\gamma} \Big( \rho(\bm{y},\bm{z}^{(2)}) - \rho(\bm{y},\bm{z}^{(2)}) \Big) \sigma^2(\bm{z}^{(1)})\sigma(\bm{z}^{(2)})\sigma(\bm{y})\\
&= 0.
\end{split}
\end{align*}

\bibliographystyle{IEEEtranN}
\bibliography{StaPLR_accepted} 


\end{document}